\documentclass[journal]{journal}
\usepackage{multirow}
\usepackage{hyperref}

\ifCLASSINFOpdf
  \usepackage[pdftex]{graphicx}
\else
\fi
\usepackage[none]{hyphenat}
\usepackage[justification=centering]{caption}	
\usepackage[labelsep=space]{caption}		
\usepackage[font=footnotesize]{caption}
\begin{document}
%
\title{JaCoText: A Pretrained Model for Java Code-Text Generation}

\author{
         Jessica López Espejel,
         Mahaman Sanoussi Yahaya Alassan,
         Walid Dahhane,
        El Hassane Ettifouri
\thanks{Jessica, Mahaman Sanoussi, Walid and El Hassane work for Novelis Research and Innovation Lab,  207 Rue de Bercy, 75012 Paris, France (e-mail: jlopezespejel@novelis.io, syahaya@novelis.io, wdahhane@novelis.io, eettifouri@novelis.io).}
}

%
%

\markboth{Journal of \LaTeX\ Class Files,~Vol.~6, No.~1, January~2007}%
{Shell \MakeLowercase{\textit{et al.}}: Bare Demo of IEEEtran.cls for Journals}
%



\maketitle
\thispagestyle{empty}

\begin{abstract}
    Pretrained transformer-based models have shown high performance in natural language generation task. However, a new wave of interest has surged: automatic programming language generation. This task consists of translating natural language instructions to a programming code. Despite the fact that well-known pretrained models on language generation have achieved good performance in learning programming languages, effort is still needed in automatic code generation.     
    In this paper, we introduce JaCoText, a model based on Transformers neural network. It aims to generate java source code from natural language text. JaCoText leverages advantages of both natural language and code generation models.  More specifically, we study some findings from the state of the art and use them to (1) initialize our model from powerful pretrained models, (2) explore additional pretraining on our java dataset, (3) carry out experiments combining the unimodal and bimodal data in the training, and (4) scale the input and output length during the fine-tuning of the model. Conducted experiments on CONCODE dataset show that JaCoText achieves new state-of-the-art results.      

\end{abstract}


%
\IEEEpeerreviewmaketitle

\section{Introduction}
%
%
%
%
    \IEEEPARstart{W}{h}en developing software, programmers use both natural language (NL) and programming language (PL). While the latter is the core component of every project, natural language is used to write documentation (ex: JavaDoc) to describe different classes, methods and variables. Documentation is usually written by experts and aims to provide a comprehensive explanation of the source code to every person who wants to use/develop the project.
    
    In the last years, the automation of programming code generation from  natural language has been studied using various techniques~\cite{ahmad2021_plbart, Phan2021_CoTexT, Guo2021_GraphCodeBERT, Xu2022_PolyCoder} of artificial intelligence (AI). Leveraging AI increases programmers productivity because it helps them automatically generate code for simple tasks, while allowing them to tackle only the most difficult ones. 

    After the big success of Transformers Neural Network~\cite{Vaswani2017_transformers}, it has been adapted to many Natural Language Processing (NLP) tasks such as question answering~\cite{devlin2018_bert, Khashabi_MultiRC2018, clark2019_boolq}, text translation~\cite{Yasmin2022_Translation} and automatic summarization~\cite{zhang2019_pegasus, Liu2019_SummAEZA}. Some of the most popular models are GPT~\cite{Radford2018_ImprovingLU, Radford2019_GPT2}, BERT \cite{devlin2018_bert},  BART~\cite{ahmad2021_plbart}, and T5~\cite{Colin2020_T5}. One of the main factors of success of these models is that they were trained on very large corpora. 
    Recently, there has been an increasing interest in programming code generation. Therefore, the scientific community based its research on proposing systems that are based on pretrained transformers.  For instance, CodeGPT and GPT-adapted~\cite{shuai2021_codexglue} are based on GPT2~\cite{Radford2019_GPT2}, PLBART~\cite{ahmad2021_plbart} is based on BART, and CoTexT~\cite{Phan2021_CoTexT} follows T5. Note that these models have been pretrained on bimodal data (containing both PL and NL) and on unimodal data (containing only PL).

\begin{figure}[!h]
        \centering
        \includegraphics[width=0.4\textwidth]{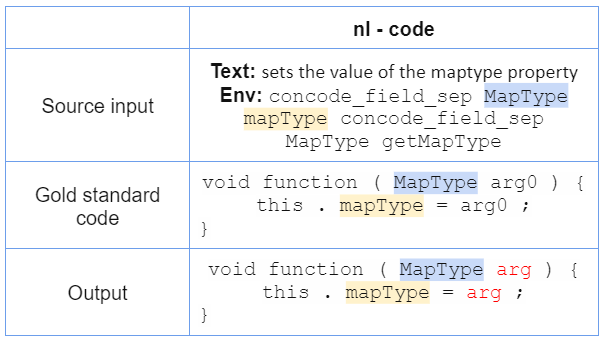}
        \caption{Example of a code generated by our model in comparison with the corresponding gold standard code}
        \label{fig:nl_code}
    \end{figure}

    
    Programming language generation is more challenging than standard text generation. This is because PLs contain stricter grammar~\cite{rabinovich2017_abstractSyntax} and syntactic~\cite{Yin2017_SyntacticNeuralModel} rules. Fig.~\ref{fig:nl_code} shows an example of an input sequence received by our model (in NL), the output of the model  (in PL) and the target code (also called gold standard or reference code). 

    In this paper, we present JaCoText,  a pretrained model based on Transformers~\cite{Vaswani2017_transformers}. First, we initialize our model from pretrained weights of CoTexT-1CC and CoTexT-2CC, instead of performing a training from scratch. Later, we conduct an additional pretraining step using data that belongs to a specific programming language (Java in our case). Moreover, unlike works that based their pretraining on CodeSearchNet~\cite{Hamel2019_CodeSearchNet} such as CodeBERT \cite{feng2020_codebert} and CoTexT \cite{Phan2021_CoTexT}, we use more java data in the pretraining stage of our model, as \cite{Radford2019_GPT2} and \cite{Colin2020_T5} have shown that Transformers neural network improves its performance significantly from increasing the amount of pretraining data. Furthermore, we carry out experiments to measure the impact of the input and output sequences length on code generation task.
    Finally, we test the unimodal data and study its impact on the model's performance. This study is crucial to evaluate the model in the pretraining stage.

     We highlight our main findings in the state-of-the-art below: 

    \begin{itemize}
        \item
        T5 has shown the best performance in language generation tasks.
        \item
        Models initialized from previous pretrained weights achieve better performance than models trained from scratch~\cite{shuai2021_codexglue, Phan2021_CoTexT}. 
        \item 
        Models such as SciBERT~\cite{beltagy2019_scibert}, and BioBERT~\cite{Lee2020_BioBERT} have shown the benefits to pretrain a model using data related to a specific domain.
        \item
        Increased data implies better training performance ~\cite{Radford2019_GPT2, Colin2020_T5}. This finding is intuitive since a large and diversified dataset helps improving the model's representation. 
        \item
        The input and output sequence length used to train the model matters in the performance of the model~\cite{Iz2020_Longformer}. 
        \item
        The objective learning used during the pretraining stage gives the model some benefits when learning the downstream tasks~\cite{Colin2020_T5, song2019_mass, Liello2021_EfficientPO}. 
    \end{itemize}

\section{JaCoText}

    In this section, we describe the core component of JaCoText to achieve state-of-the art results. 
    
    \subsection{Fine-tuning}
    
    We fine-tune our models based on two criteria:
    
    \paragraph{Sequence Length} After analyzing in detail the outputs generated by previous works, we observed that some of the code sequences produced by the models were incomplete compared to the target ones. Consequently, we tokenized the training and validation sets with SentencePiece model~\cite{Kudo2018_SentencePiece}. We then computed the largest sequence data, and used its length  for both the inputs and the targets. 

    \paragraph{Number of steps} Since we increased the length of sequences in our model, we increased the number of fine-tuning steps. According to \cite{Colin2020_T5}, a way to improve the model's performance is by increasing the number of steps in the training.

    We apply both criteria initializing the fine-tuning from CoTexT checkpoints 2CC and 1CC, respectively. CoTexT-1CC is pretrained on unimodal data (only code), and CoTexT-2CC is pretrained on bimodal data (both code and natural language). Results of these experiments are shown in Table~\ref{tab:vary_steps_length_seq}.  
    
    \subsection{Additional Pretraining} 
    
    Authors of~\cite{Colin2020_T5} made some important observations that we support in our work: (1) for some specific tasks, the way to improve the model's performance is to pretrain it with a dataset that belongs to a specific domain, (2) additional pretraining can improve the performance of a model, (3) a low number of epochs when pretraining a model leads to higher scores in generation tasks. Besides, other works such as 
    CodeGPT-adapted \cite{shuai2021_codexglue} and CoTexT \cite{Phan2021_CoTexT} show that models initialized with pretrained weights achieve better results than models trained from scratch. 

\begin{figure}[!ht]
        \centering
        \includegraphics[width=0.4\textwidth]{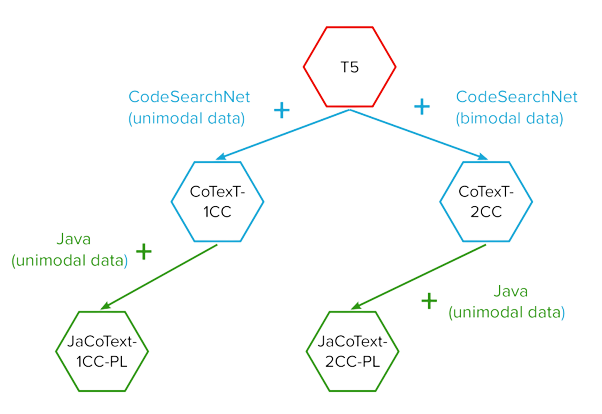}
        \caption{JaCoText model, best viewed in color}
        \label{fig:model_JaCoText}
    \end{figure}

    Based on the previous highlighted points, we carried out additional pretraining on unimodal java data (Fig.~\ref{fig:model_JaCoText}). We initialized JaCoText-B-1CC-PL and JaCoText-B-2CC-PL models from pretrained weights of CoTexT-1CC and CoTexT-2CC~\cite{Phan2021_CoTexT}, respectively. We trained the previous models on only-code sequences. We follow the same procedure for both $T5_{base}$ and $T5_{large}$. The input of the encoder is a noisy Java code. The input of the decoder is the original Java code with one position offset.

    Briefly, once the model is initialized from T5 weights (previously pretrained on C4 dataset), we further pretrain it on CodeSearchNet and our java dataset. Later, we use the final checkpoints to initialize the fine-tuning on CONCODE dataset.

\section{Experimental Setup}
\label{sec:experimental}

    \subsection{Architecture} 
    \label{subsec:Architecture}
    
    JaCoText uses the same architecture as $T5$~\cite{Colin2020_T5}, which is based on Transformers~\cite{Vaswani2017_transformers}. On the one hand, $T5_{base}$ consists of 12 layers in both the encoder and the decoder, with model dimension of 768 and 12 heads (approx. 220M parameters). On the other hand, $T5_{large}$ has 24 layers in both the encoder and the decoder, with model dimension of 1024 and 16 heads (approx. 770M parameters).

    \subsection{Code Generation Dataset}
    \label{subsec:code_generation}
    
    To perform our experiments in Java code generation task, we used CONCODE~\cite{Iyer2018_MappingLTC}, a dataset that contains context of a real world Java programming environment. CONCODE aims to generate Java member functions that have class member variables from documentation. Table~\ref{tab:concode} describes  CONCODE dataset.

    \begin{table}[h!]
    \centering
            \caption{\footnotesize \normalfont\scshape \\A Summary of CONCODE Dataset}
            \begin{tabular}{cccc} \hline\hline
        & \multicolumn{3}{c}{\textbf{Size}} \\ \hline
        \textbf{Category} & Train   & Val   & Test     \\ \hline
        \# Text-Code lines   & 100K       & 2K       & 2K  \\ \hline \hline       
        \end{tabular}
        
        \label{tab:concode}   
    \end{table}

    \vspace{-1.5em}

    \subsection{Additional Pretraining Dataset}
    
    For the additional pretraining, we used our Java dataset. Originally, it consists of $812,008$; $40,468$, and $51,210$ samples in the training, validation, and test sets, respectively. We deleted the problematic samples in the three sets ($2974$ in the training set, $235$ in the validation set, and $161$ in the test set). We use the rest of samples ($900,316$) from the three sets to pretrain our model. 

    \subsection{Evaluation Metrics}
    \label{subsec: evaluation_metrics}
    
    To evaluate our models, we used the three metrics described below. 
        
        \paragraph{BLEU}~\cite{papineni2002_bleu} is a metric based on n-gram precision computed between the candidate and the reference(s). N-gram precision penalizes the model if: (1) there are words that appear in the candidate but not in any of the references, or (2) if a word appears more times in the candidate than in the maximum reference count. However, the metric fails if the candidate does not have the appropriate length. 
        Following~\cite{ahmad2021_plbart} and~\cite{Phan2021_CoTexT} we use the corpus-level BLEU score in the code generation task. 
        
        \paragraph{CodeBLEU}~\cite{Ren2020_CodeBLEUAM}  works via n-gram match, and it takes into account both the syntactic and semantic matches. The syntax match is obtained by matching between the code candidate and code reference(s) sub-trees of abstract syntax tree (AST). The semantic match considers the data-flow structure.
        
        \paragraph{Exact Match (EM)} is the ratio of the number of predictions that match exactly any of the code reference(s).
        
    \subsection{Baselines}
    \label{subsec:baselines}
    
    We compare our model with four state-of-the-art Transformer-based models. 
    
    \begin{itemize}
        
        \item CodeGPT, CodeGPT-adapted~\cite{shuai2021_codexglue} are based on GPT-2 model~\cite{Radford2019_GPT2}. The difference between both models is that CodeGPT is trained from scratch on CodeSearchNet dataset~\cite{Hamel2019_CodeSearchNet}, while CodeGPT-adapted is initialized from GPT-2 pretrained weights. 
        
        \item PLBART~\cite{ahmad2021_plbart} uses the same architecture than $BART_{base}$~\cite{lewis2020_bart}. Additionally, PLBART uses three noising strategies: token masking, token deletion and token infilling. 
    
        \item CoTexT~\cite{Phan2021_CoTexT} uses the same architecture than $T5_{base}$. It is trained on both unimodal and bimodal data using CodeSearchNet Corpus~\cite{Hamel2019_CodeSearchNet}, and GitHub Repositories.
        
    \end{itemize}

\section{Results and Discussion}
\label{sec:results}

    Firstly, we study the performance of T5 model on the Java generation task. We directly fine-tune on CONCODE dataset three types of T5: $T5_{base}$, $T5_{large}$, and $T5_{3B}$. The best parameters we used are highlighted in Table~\ref{tab:vary_steps_length_seq}. 
    
    \begin{table}[!hbt]
    \centering 
 \caption{\footnotesize \normalfont\scshape \\Results Obtained When Fine-Tuning Directly from T5 Models}
    \begin{tabular}{cccc}
    \hline \hline
    \multicolumn{1}{c}{\textbf{Parameter}} & \multicolumn{3}{c}{\textbf{Metrics}}           \\ \hline
    \textbf{\# steps}                        & \textbf{BLEU} & \textbf{EM} & \textbf{CodeBLEU} \\ \hline
    \multicolumn{4}{c}{\textbf{T5-base}}                                         \\ \hline
    45000   & 34.03 & 20.45  & 36.73   \\
    60000   & 34.08  & 20.30 & 37.00    \\ \hline
    \multicolumn{4}{c}{\textbf{T5-Large}}                                       \\ \hline
    45000   &  34.00 &  20.30  & 36.98  \\
    60000   &  36.23 &  21.05  & 38.84  \\ \hline
    \multicolumn{4}{c}{\textbf{T5-3B}}                                                         \\ \hline
    45000    &  32.65  &  21.60  &  35.47   \\
    60000    &  35.68  &  21.65  &  38.37  \\
    90000    &  36.28  &  22.50  &  38.97    \\
    120000   &  \textbf{38.11}  &  \textbf{22.20}  &  \textbf{40.81}   \\  \hline \hline
    Best results are in bold.
    \end{tabular}

    \label{tab:fine_tuning_from_T5_directly}
    \end{table}
    
    Table~\ref{tab:fine_tuning_from_T5_directly} provides the scores of each type of T5 models directly after the fine-tuning using CONCODE dataset. In all cases, the score improves as the number of steps increases. Unsurprisingly, the most sophisticated  $T5_{3B}$ model gets the best results, followed by $T5_{large}$ and $T5_{base}$, while $T5_{3B}$ takes more time to converge.

    Table~\ref{tab:vary_steps_length_seq} provides results obtained when varying the number of steps and the length of input and output sequences while fine tuning CoTexT-2CC and CoTexT-1CC checkpoints on CONCODE dataset. 
    Results show that using $60000$ steps provides better results than using $45000$ steps in the fine-tuning as noted in~\cite{Phan2021_CoTexT}. In addition, by using the largest code sequence length, we outperform the BLEU and EM scores obtained by~\cite{Phan2021_CoTexT} (highlighted in italic).  Results vary slightly, almost undetectable. However, CoTexT-1CC performs better using BLEU and CodeBLEU, while CoTexT-2CC achieves better results using the EM metric.

    \begin{table*}[!h]
        \centering
        \caption{\footnotesize \normalfont\scshape \\Results When Varying the Number of Steps, and the Input and Output Sequence Length during CoTexT-2CC and CoTexT-1CC Fine-Tuning}
        \begin{tabular}{ccccc}
        \hline \hline
        \multicolumn{2}{c}{\textbf{Parameters}}                                                & \multicolumn{3}{c}{\textbf{CoTexT 2CC / 1CC}}                                               \\ \hline
        \textbf{\begin{tabular}[c]{@{}c@{}}input / target \\ length\end{tabular}} 
        & \textbf{\# steps} & \textbf{BLEU} & \textbf{EM}   & \textbf{\begin{tabular}[c]{@{}c@{}}Code\\ BLEU\end{tabular}} \\ \hline
            256 / 256  & 45000   & \textit{36.51 / 37.40} & \textit{20.10 / 20.10} & \textit{39.49 / \textbf{40.14}} \\
            256 / 256  & 60000   & 36.22 / 36.00 & 20.85 / 20.20 & 38.88 / 38.62 \\
            256 / 379  & 60000   & 36.60 / 36.45 & 20.10 / 20.10 & 39.23 / 39.20  \\
            379 / 379  & 45000   & 37.08 / 37.33  & 21.50 / 21.25 &  39.80 / 39.85  \\
            \textbf{379 / 379}  & \textbf{60000}   & 37.46 / \textbf{37.66} & \textbf{21.45} / 21.40 & 39.94 / 40.03  \\
            200 / 200  & 45000   & 34.61 / 34.79  & 19.65 / 19.30  & 37.64 / 37.60  \\
            200 / 200  & 60000   & 35.17 / 35.23 & 19.10 / 18.20 & 38.12 / 38.19    \\ \hline \hline                                     
        \end{tabular}
        
        \label{tab:vary_steps_length_seq}
    \end{table*}

     Varying the number of steps and augmenting the length of both the input and target in the fine-tuning provide the first step to improve results on Java code generation task. The second step consists in using the additional pretraining from CoTexT weights following \cite{shuai2021_codexglue}. After additional pretraining, we fine-tune the model using the best parameter values from Table~\ref{tab:vary_steps_length_seq}. 
    
     Table~\ref{tab:results_addional_pretraining} provides fine-tuning results after performing the additional pretraining using our Java dataset. The models are initialized with JaCoText-B weights when they are trained following the $T5_{base}$ architecture, and with JaCoText-L weights when they are trained following $T5_{large}$. As we mentioned previously, the additional training using our Java dataset is initialized from CoTexT weights. However, the training of JaCoText-L-1CC-PL and JaCoText-L-2CC-PL models started from $T5_{large}$ weights (previously trained on C4~\cite{Colin2020_T5} dataset). We trained $T5_{large}$ on CodeSearchNet dataset, and later on our Java dataset during $200,000$ steps each and using unimodal data (PL only). Finally, we fine-tune the model on CONCODE dataset for $45,000$ steps.

    Results show that JaCoText achieves state-of-the-art results. Unsurprisingly, JaCoText-L models get the highest scores using the three metrics, because $T5_{large}$ has a more sophisticated architecture. In addition, it is noteworthy to mention that in both architectures, \textit{base} and \textit{large}, the best results are obtained with models that were pretrained on bimodal data. This finding proves that  training models with bimodal data performs better than with unimodal data.         
    
    \begin{table}[h!]
    \centering
        \caption{\footnotesize \normalfont\scshape \\Results with Additional Pretraining Using our Java Dataset}
        \begin{tabular}{c c c c}
        \hline  \hline        
        \textbf{Model}                & \multicolumn{1}{c }{BLEU} & \multicolumn{1}{c }{EM} & \multicolumn{1}{c}{CodeBLEU} \\ \hline
        \cite{Guo2019_Coupling} & 24.40 & 10.05 & 29.46 \\ 
        CodeGPT       & 28.69   & 18.25   & 35.52   \\ 
        CodeGPT-Adp   & 32.79   & 20.10   & 35.98    \\
         PLBART       & 36.69   & 18.75   & 38.52   \\
        T5-base       & 32.74   & 18.65   & 35.95   \\
        CoTexT-2CC    & 36.51   & 20.10  & 39.49   \\
        CoTexT-1CC    & 37.40   & 20.10   & 40.14   \\ \hline
        \small JaCoText-B-1CC-PL   & 38.65   & 21.85   & 41.19    \\
        \small JaCoText-B-2CC-PL   & \textbf{39.07}   & \textbf{22.15}  & \textbf{41.53}  \\ \hline
        \small JaCoText-L-1CC-PL   & 39.67   & 22.30  & 42.19  \\ 
       \small  JaCoText-L-2CC-PL   & \textbf{39.87}      & \textbf{22.45}    & \textbf{42.49}   \\ \hline  \hline
       Best results are in bold.
        \end{tabular}
   
        \label{tab:results_addional_pretraining}
            \end{table}
    
\begin{figure}[!h]
        \centering
        \includegraphics[width=0.35\textwidth]{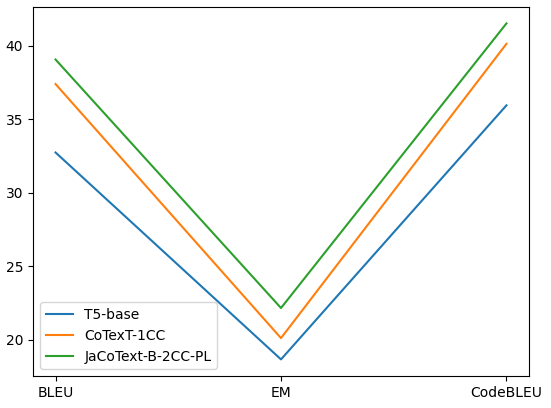}
        \caption{Improvement of our model through the additional training}
        \label{fig:improvements}
    \end{figure}

    Finally, Fig.~\ref{fig:improvements} shows the improvements of our model JaCoText-B-2CC-PL with an additional training using our Java dataset. For a fair comparison, the three models are fine-tuned for $60,000$ steps, and they all follow the $T5_{base}$ architecture. 

\section{Related Work}
\label{sec:related_work}
    
    Early interesting approaches mapped natural language to source code using regular expressions~\cite{locascio2016_neuralRegularExpr} and database queries~\cite{Xu2017_queriesFromNL,Zhong2017_Seq2SQLGS}. Most recently, neural networks have proven their effectiveness to automatically generate source code from different general-purpose programming languages like Python~\cite{Yin2017_SyntacticNeuralModel} and Java~\cite{Phan2021_CoTexT}. Simultaneously, large-scale datasets have surged in order to facilitate tackling the problem. These datasets include  CONCODE~\cite{Iyer2018_MappingLTC}, CONALA~\cite{Yin2018_conala},  and CodeSearchNet~\cite{Hamel2019_CodeSearchNet}. 
    
   Reference \cite{Yin2017_SyntacticNeuralModel} used a BiLSTM encoder, and an RNN decoder to generate syntactically valid parse trees. Inspired by the grammar-aware decoder, \cite{Iyer2018_MappingLTC} used Bi-LSTMs encoder to compute the contextual representations of the NL, and an LSTM-based RNN decoder with two-step attention mechanism followed by a copying mechanism to map NL with the source code. 

    Recently, models based on Transformers~\cite{Vaswani2017_transformers} and originally intended for the generation of natural language have been of a great benefit for automatic code generation. PLBART uses the same model architecture as $BART_{base}$~\cite{lewis2020_bart}. Unlike $BART_{base}$, PLBART  stabilizes the training by adding a normalization layer on the top of both the encoder and the decoder, following~\cite{Liu2020_MultilingualDP}. Similarly to PLBART, CoTexT (Code and Text Transfer Transformer)~\cite{Phan2021_CoTexT} is an encoder-decoder model, and it follows $T5_{base}$~\cite{Colin2020_T5} architecture. 
    
    Moreover, encoder-only models such as RoBERTa-(code) \cite{shuai2021_codexglue} inspired by RoBERTa \cite{liu2019_roberta}, and decoder-only models like CodeGPT and CodeGPT-adapted have achieved competitive results in the state of the art. 
    Similarly to CodeGPT and CodeGPT-adapted, RoBERTa-(code) is  pretrained on CodeSearchNet dataset. Unlike RoBERTa-(code), CodeGPT is pretrained on CodeSearchNet from scratch, and CodeGPT-adapted is pretrained starting from pretrained weights of GPT-2~\cite{Radford2019_GPT2}. Both CodeGPT and CodeGPT-adapted follow the same architecture and training objective of GPT-2.

\section{Conclusion}
\label{sec:conclusion}

    We present JaCoText, a set of T5-based~\cite{Colin2020_T5} pretrained models designed to generate Java code from natural language. We evaluate the performance of three architectures: $T5_{base}$, $T5_{large}$, and $T5_{3B}$ to generate Java code. We follow the recommendations proposed by~\cite{Phan2021_CoTexT, Colin2020_T5, shuai2021_codexglue} to improve the performance of T5 model on Java code generation. Some takeaways from these experiments are: (1) pretraining the model using a dataset designed to tackle a specific task is beneficial, (2) additional pretraining can improve the performance of the model, and (3) using a low number of epochs in the pretraining helps improving the final performance.

    Our models achieve state-of-the-art results on the Java code generation task. We prove that, each modification in our models,   such as the additional training, allows JaCoText to have better comprehension of the java programming language. In the future, it would be interesting to explore other neural network models performance, and improve the programming language syntax through the decoding algorithm. In addition, since in this paper we focus our work on additional training using code only, we leave additional training using bimodal data for future work.







%


\vspace{-5em}

\begin{biography}[{\includegraphics[width=1in,height=1.25in,clip,keepaspectratio]{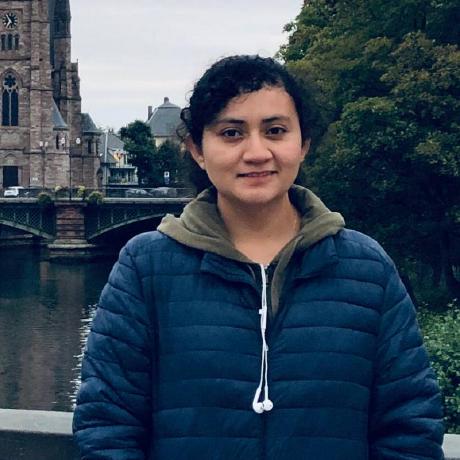}}]\\ \textbf{Jessica López Espejel} is a deep learning researcher at Novelis Research and Innovation Lab. Her research is focused on Automatic Code Generation and Transformers Neural Networks. She holds a Ph.D. in Natural Language Processing from Sorbonne Paris Nord University and CEA-LIST  (2021). Email: $jlopezespejel@novelis.io$.
\end{biography}

\vspace{-5em}

\begin{biography}[{\includegraphics[width=1in,height=1.25in,clip,keepaspectratio]{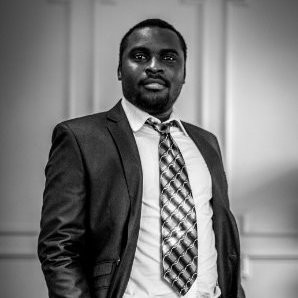}}]\\ \textbf{Mahaman Sanoussi Yahaya Alassan} works as a researcher at Novelis Research and Innovation Lab.  His research focuses on semantic text classification, information retrieval, named entity extraction. He obtained his Ph.D. in NLP from Paris Nanterre University in 2017.
Email: $syahaya@novelis.io$
\end{biography}

\vspace{-3em}

\begin{biography}[{\includegraphics[width=1in,height=1.25in,clip,keepaspectratio]{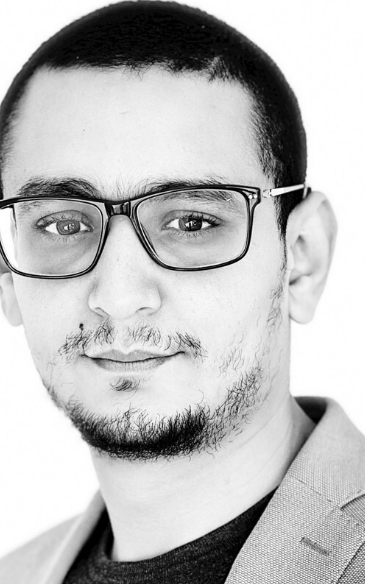}}]{Walid Dahhane} is the CTO and Co-founder of Novelis. He is an Entreprise architect, a specialist in microservices and Smart Automation architectures, and a doctor in AI \& NLP. He is in charge of the IS Urbanisation and Cybersecurity and manages the activities around the business solutions. Email: $wdahhane@novelis.io$
\end{biography}

\vspace{-2em}

\begin{biography}[{\includegraphics[width=1in,height=1.25in,clip,keepaspectratio]{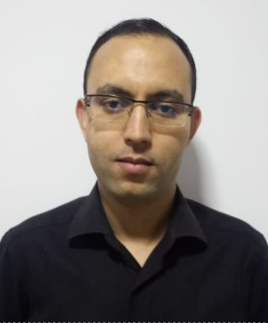}}]\\ \textbf{El Hassane Ettifouri} holds a Ph.D. in software engineering and Artificial Intelligence. He is an Associate and the Head of Novelis Research and Innovation Lab. His research focuses on Artificial Intelligence, Natural Language Processing, and Computer Vision.
He was professor in ENSAO and SupMTI engineering schools, and was also the founder of the ZeroCouplage Framework. Email: $eettifouri@novelis.io$
\end{biography}

\vfill








\end{document}